\documentclass[runningheads]{llncs}

\usepackage{amsmath}
\usepackage{multirow}
\usepackage{graphicx}
\usepackage{cite}
\begin{document}
\title{An External Knowledge Enhanced Multi-label Charge Prediction Approach with Label Number Learning}
\titlerunning{Charge Prediction with Label Number Learning}
%
\author{Duan Wei \and Li Lin}

\authorrunning{W.Duan et al.}
%
\institute{School of Computer Science and Technology, Wuhan University of Technology, Wuhan 430070, China\\
\email{314585797@qq.com, cathylilin@whut.edu.cn}}
%
\maketitle              
\begin{abstract}
Multi-label charge prediction is a task to predict the corresponding accusations for legal cases, and recently becomes a hot topic. However, current studies use rough methods to deal with the label number. These methods manually set parameters to select label numbers, which has an effect in final prediction quality. We propose an external knowledge enhanced multi-label charge prediction approach that has two phases. One is charge label prediction phase with external knowledge from law provisions, the other one is number learning phase with a number learning network (NLN) designed. Our approach enhanced by external knowledge can automatically adjust the threshold to get label number of law cases. It combines the output probabilities of samples and their corresponding label numbers to get final prediction results. In experiments, our approach is connected to some state-of-the art deep learning models. By testing on the biggest published Chinese law dataset, we find that our approach has improvements on these models. We future conduct experiments on multi-label samples from the dataset. In items of macro-F1, the improvement of baselines with our approach is 3\%-5\%; In items of micro-F1, the significant improvement of our approach is 5\%-15\%. The experiment results show the effectiveness our approach for multi-label charge prediction.

\end{abstract}
\section{Introductionn}
In recent years, NLP (Nature Language Processing) has a huge development in many research tasks, such as text classification, NER (Named Entity Recognition), semantic labeling, reading comprehending, etc. With the development of internet, the unit of global data total has crossed from GB to ZB. Among them, various valuable text data has greatly promoted the development of the NLP field, and help to alleviate defects of these works.

The legal charge prediction is a multi-label text classification task that learns from law case called fact to predict accusations (labels). This task makes use of the sample's complex text data to properly classify it into the appropriate labels (accusation types). A sample have one label or multiple labels, and there are subtle differences of contents of texts. For example, someone has violated two laws of theft and robbery because of a series of criminal activities. Therefore, we need to understand the description of a case called fact, extract the description of criminals breaking the law, and finally classify the sample into the corresponding accusations. In public Chinese law datasets, like Cail2018, the number of single-label samples is the largest. The number of training data is 154,592, but there is a huge difference between the number of single-label samples and multi-label samples. The number of multi-label samples accounts for a small portion of the total. The more labels the samples have, the less the size of them is in the total amount. For example, the number of 5 labels samples is 58 and the number of 9 labels sample is 1. Therefore multi-label legal charge prediction is a difficult multi-label text classification prediction task.

There are many models that can better solve text classification. For example, Kim purposed the TextCNN~\cite{paper_1} that can synthesize local text content of different receptive field to capture the most important features of text; Shi purposed the CRNN~\cite{paper_2} that  can capture the local information, global information, and interrelated associations of texts; Johnson purposed the DPCNN~\cite{paper_3} that can get more efficient and widely available global information while effectively obtaining local information; Vaswani purposed Multi-head Attention that can find out the degree of correlation between the text and the label~\cite{paper_13}; And there are GRU and LSTM~\cite{paper_4,paper_5} that can get better capture of longer distance text dependencies. We argue that for the legal charge prediction task, these models will be improved by the external knowledge from law provisions and a suitable way to decide label numbers.

Firstly, all models are only learning the logic between the content of the case and accusations, but they ignore the correlation between the content of the case and law provisions of corresponding accusations. If the model already has such a priori knowledge—the law provisions, it is easy to learn which case is more relevant to the accusations and assist itself learning to understand the content of the case, and finally make the prediction result more accurate.

Secondly, their models only get the output probability of belonging to the corresponding label. However, there is only two widely used strategies to map the probability to output label number as follow: 

(1) Top-k strategy: For the output probability of each sample, it selects the first k labels that has the highest probability as the final prediction result;

(2) Threshold strategy: For each probability of the sample, it selects a value as the threshold, and all labels with the probability greater than the threshold are the final prediction result.

However, these two strategies have some disadvantages: 

(1) We need to calculate the distribution, maximum value, median of the entire output probability. After getting understanding of the data, we can adjust the parameters to obtain the optimal result. However, it will take a lot of time and efforts to complete this goal;

(2) or most models or datasets, their output probabilities must be different. Therefore, we should select a threshold or k value respectively. It also takes time and efforts; 

(3) These two widely used strategies will cause certain errors no matter how the parameters are selected, and the errors will greatly increase for the multi-label task.

Therefore, we propose an external knowledge enhanced multi-label charge prediction approach with automatic label number learning. Meanwhile, we purpose an efficient network structure called number learning network (NLN). For this legal text classification problem, we get external knowledge-corresponding to the law provision of each accusation, and assisting various deep learning models to understand legal texts. And the output probability can be mapped to probability of the label number through NLN. At last, combining them can get the better result than using popular deep learning models alone. Among it, the most important technology of the network is to construct a special embedding layer, that adaptively adjusts the threshold of each corresponding label probability through BP. After implementing this network any deep learning models, it is still an end-to-end model and can effectively solve the above problems..

In this paper, we have an overview of the legal text processing, multi-label text classification, memory mechanism and label number learning in Section 2. We propose an external knowledge enhanced end-to-end multi-label charge prediction approach with automatic label number learning and a number learning network (NLN). This network can be implemented with various deep learning model for the multi-label legal charge prediction. The details of approach will be elaborated in the Section 3. For the experiment results, we use the six deep learning methods with NLN on multi-label samples and our approach on the dataset. This approach has a huge improvement on these models, and the promotion is significant in the multi-label samples. The experiment results show that adding label number learning can also improve the model effect. The more details of experiment will be elaborated in the Section 4. Section 5 makes a conclusion of our work and discuss our future work.

\section{Related Work}
\subsection{Legal Text}

As the development of deep learning technology, there are many fields of NLP applications. Because of legal text form and the large amount of data, the law field of NLP becomes a hot topic. Luo adopted the attention based neural network with the supplement of relevant law articles~\cite{paper_6,paper_107,paper_108}; Some researchers thought decisions of applicable law articles, charges, fines, and the terms of penalty have logical relationship~\cite{paper_7}; Some researchers attempted a hybrid approach to summarize legal case reports~\cite{paper_8}; And some researchers compared policy differences by the embedding of entire legal corpus of each institution~\cite{paper_9}. More and more researchers paid attentions on the legal text, and exploited lots of deep learning models.~\cite{paper10, paper11, paper12}. However, for the classification of legal text, the current research is far away from real application.

\subsection{Multi-Label Text Classification}

In the direction of text classification, so many models have been proposed, and have achieved quite good results that makes an outstanding contribution to the development of this field. Kim purposed the attention-based classifier that can achieve multi-label emotion classification~\cite{paper_10,paper_103}; Yang applied a sequence generation model with a novel decoder structure to solve it with correlations between labels~\cite{paper_11,paper_104,paper_105}; ~\cite{paper_12,paper_13,paper1,paper_14} created various attention mechanisms for NLP that applied into text classification by others; There was a new module that can sheep up training~\cite{paper_15,paper_16}; Zhang achieved text classification with the correlation between different task data~\cite{paper_17}; And~\cite{paper_18} explored the influence of different semantic embedding to multi-label text classification. And various purposed embedding approaches for representation in document can be applied in multi-label text classification~\cite{paper2,paper3,paper4,paper5,paper_106}. Most researchers put their energy into complex or novelty deep learning model to solve this problem, however, few researchers refer to the label number learning.

\subsection{Memory Mechanism}

The existing model has a weak storage capacity for memory, which can’t store too much information, and it is easy to lose some semantic information, so the memory network memorizes information by external storage. Facebook AI proposed a memory mechanism to enhance memory of network, and later solved the problem of not being able to end-to-end train~\cite{paper_100}. In the next study, it enhanced the size of memory and proposed the new dynamic memory network structure~\cite{paper_101,paper_102}. Therefore, we got the external knowledge through this memory mechanism to enhance the classification ability of the model.

\subsection{Label Number Learning} 
 
 Label number learning is to learning the label number on the basis of predicted probability of deep learning model. There are two common strategies to solve this problem:1. Select an appropriate global threshold. If the predicted probability greater than it, the label will be true label; 2. Directly determine the number of labels of sample, such as top-k. However, there are respectively some defects. Yang purposed the approach that summarizes the whole process of multi-label classification. And she thought there is a need to learning to optimize the threshold over ranked listed on the per-instance basis~\cite{paper_19}; Lenc refered to a simple neural network that can solve the problem by top-k~\cite{paper_20}. Then these papers give the idea whether exist a neural network that can learn the label number. If it exists, then we can add this network after any deep learning model of the general prediction approach to enhance the performance of the model.

\section{Our Approach}
\subsection{Problem Definition}

The charge prediction is a task to predict accusations by analyzing the case description called fact in the Cail2018. In Figure 1 we see that a fact in law is a series of precise descriptions of the case and the presentation of criminal behavior. After segmentation of fact, we define the input as $T_i=\{t_{i1},t_{i2},t_{i3},\ldots,t_{ij}\}$, , and each $t_{ij}$ is a word, such as ‘ability’. In Figure 1, we analyze the facts through deep learning model, find out the text of the most relevant accusations (the text part on the yellow background), and then understand their association with each accusation, and finally predict that accusations of sample are respectively arson and intentional injury. We want to get the output is $R_i$, each $R_i$ is a list, such as  $R_i=[1,55,120]$. Each number in the list represents the index of the corresponding accusation, such as arson or intentional injury. And we purpose our own approach and NLN on the problem definition.

\begin{figure}
	\includegraphics[width=\textwidth]{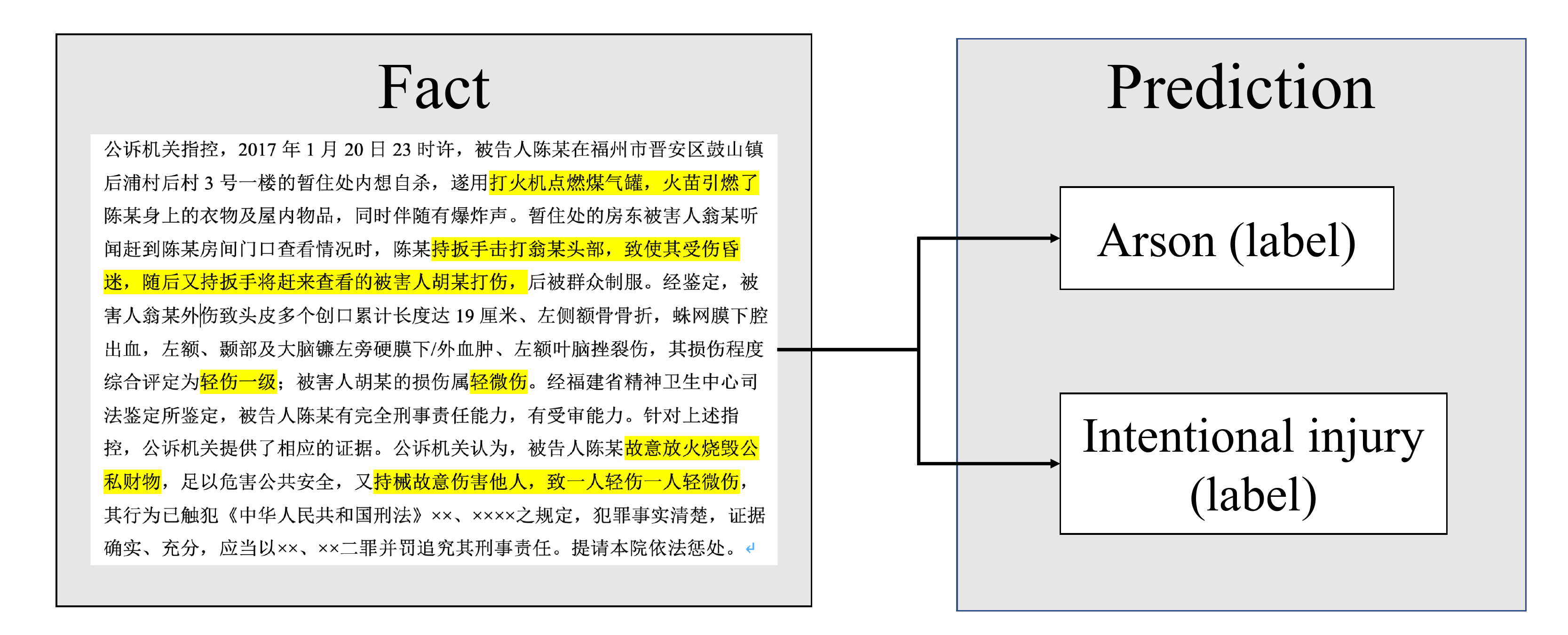}
	\caption{Example of charge prediction} \label{fig1}
\end{figure}

\subsection{ The Framework of Our Approach}
For multi-label text classification tasks, our purposed prediction approach is: (1) use text preprocessing to obtain text vectors of input data and external knowledge; (2) use deep learning model and attention mechanism to obtain vectors, and combine them to get output results; (3) By machine learning training with output probability of the case sample, the probability of the label number of each sample is obtained; (4) The final result is obtained by label decision that combines the predicted label number of the case sample with output probability of each case sample. This new approach with automatic label number learning is also end-to-end. For the third step of the approach, we can think of it as an auxiliary system. And details of the framework of our approach are shown in Figure 2.

In this approach, we can use any text vectorization methods, whether it is one-hot, word-embedding or direct learning, etc. You can also use any kinds of deep learning models, whether it is TextCNN, Bi-GRU, etc. Therefore, it is easy to experiment for the charge prediction task. We make use of the framework of the memory network to obtain the correlation between text and legal provisions by attention mechanism in training, and combine their outputs
\begin{figure}
	\includegraphics[width=\textwidth]{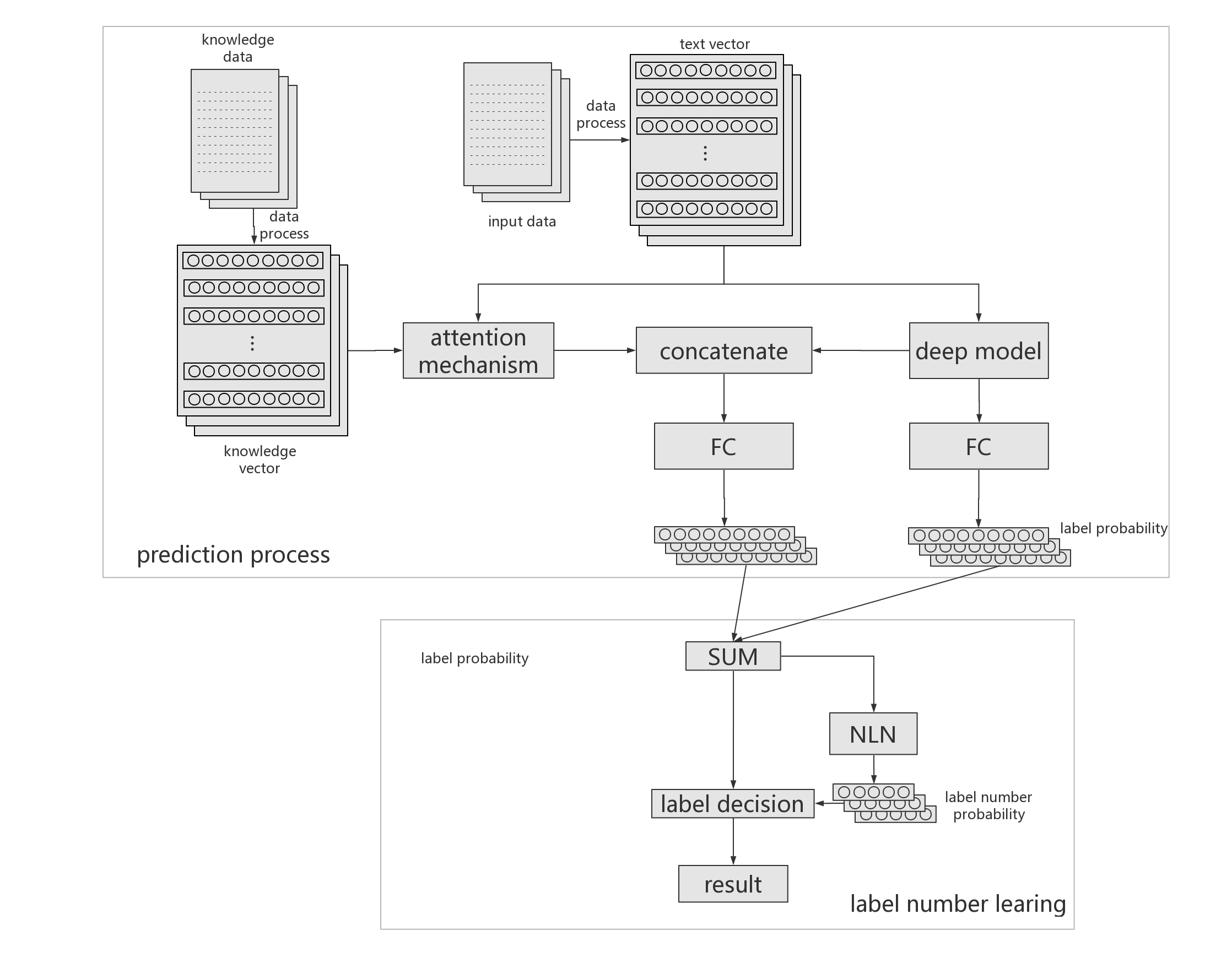}
	\caption{External knowledge enhanced prediction with label number learning} \label{fig2}
\end{figure}
 with the joint training to get the results. Since both outputs are obtained at the same time, the loss function also needs to be adjusted. See 3.2 for details. Through this approach, adding a label number learning phase after the general prediction approach can effectively solve the multi-label number decision problem. In current research, for the output of the model prediction, both of top-k or threshold strategies need to observe the actual output, and try to manually determine the parameters. However, for the multi-label task, the number of labels is variable, and the probability distributions of different labels is different, so both strategies will produce a huge judgment error. Using this approach, the number of multi-label charge prediction can be determined adaptively and the results are more accurate. 

\subsection{Loss Function}
Since we hope that the network can learn to understand the logic of the text and it can also make decisions by matching the relevance of the text to the legal provisions, the network has multiple outputs. So we set the loss of these two outputs to $loss_1$ and $loss_2$respectively, and set the $Loss$ of the network to their weighted sum by the custom weight.

\begin{equation}
Loss = w_1 * loss_1 + w_2 * loss_2
\end{equation}

$w_i$ is for them its respective weight.

\subsection{Number Learning Network (NLN)}
This paper adds a label number learning phase after general prediction process phase according to~\cite{paper_0}, and proposes a network that can better solve the label number problem of charge prediction called number learning network (NLN). Generally, the input of number learning network is $D=\{X_i,Y_i\}_{i=1}^m$. For the number learning network, $X_i$ is the label output probability obtained from the previous phrase text prediction, $Y_i$ is the corresponding one-hot encoding of number of labels. The details of $X_i$ and  $Y_i$ are illustrated in equation 1,2.

\begin{equation}
X_i=\{x_{i1},x_{i2},x_{i3},\ldots,x_{ij}\}, x_{ij}\in[0,1]
\end{equation}

\begin{equation}
Y_i=\{y_{i1},y_{i2},y_{i3},\ldots,y_{ik}\}, x_{ik}=0,1
\end{equation}

The formula for neural networks is:

\begin{equation}
x_{i+1}=f(\sum_{i=0}^{m-1}w_{ij}x_i+b_j)
\end{equation}

For the purpose of the experiment that it can adaptively adjust the threshold of the corresponding label, this paper modifies the network and purposes the number learning network (NLN). The most significant difference between the two networks is that the weight value of the first layer is fixed. For example, the first neuron of the first hidden layer is only connected to the first neuron of the second hidden layer. The specific expression is as follows:

\begin{equation}
w_{ij}=
\begin{cases}
1,  \quad i=j \\
0, \quad i\not =j 
\end{cases}
\end{equation}

Secondly, the number of hidden neurons fixed in the first hidden layer is the same as the number of neurons in the input layer. Through these two steps, this article creates a specific embedding process before the output probability is inputted into the FC network. For each $x_{ij}$, in the embedding process, it represents the probability that the sample belongs to $j$th accusation. Through this step of the embedding process, if the equality 5 is established, then $x_{ij}$ can pass the filtering.

\begin{equation}
f(x_{ij}+b_j)>0,  \quad i=j
\end{equation}

$b_j$ is the threshold corresponding to all $x_{ij}$. In this paper, the back propagation (BP) process can adaptively adjust the $b$ for each label as the corresponding threshold. There is a fine adjustment or screening work through the following several layers of fully connected (FC) network. After the specific embedding process, the FC network is used to map values of each label to the output probability. At the same time, the network can not be too complex to prevent gradient explosions or gradient dispersion. The details of network are illustrated in Figure 3.

\begin{figure}\centering
	\includegraphics[width=8cm,height=6cm]{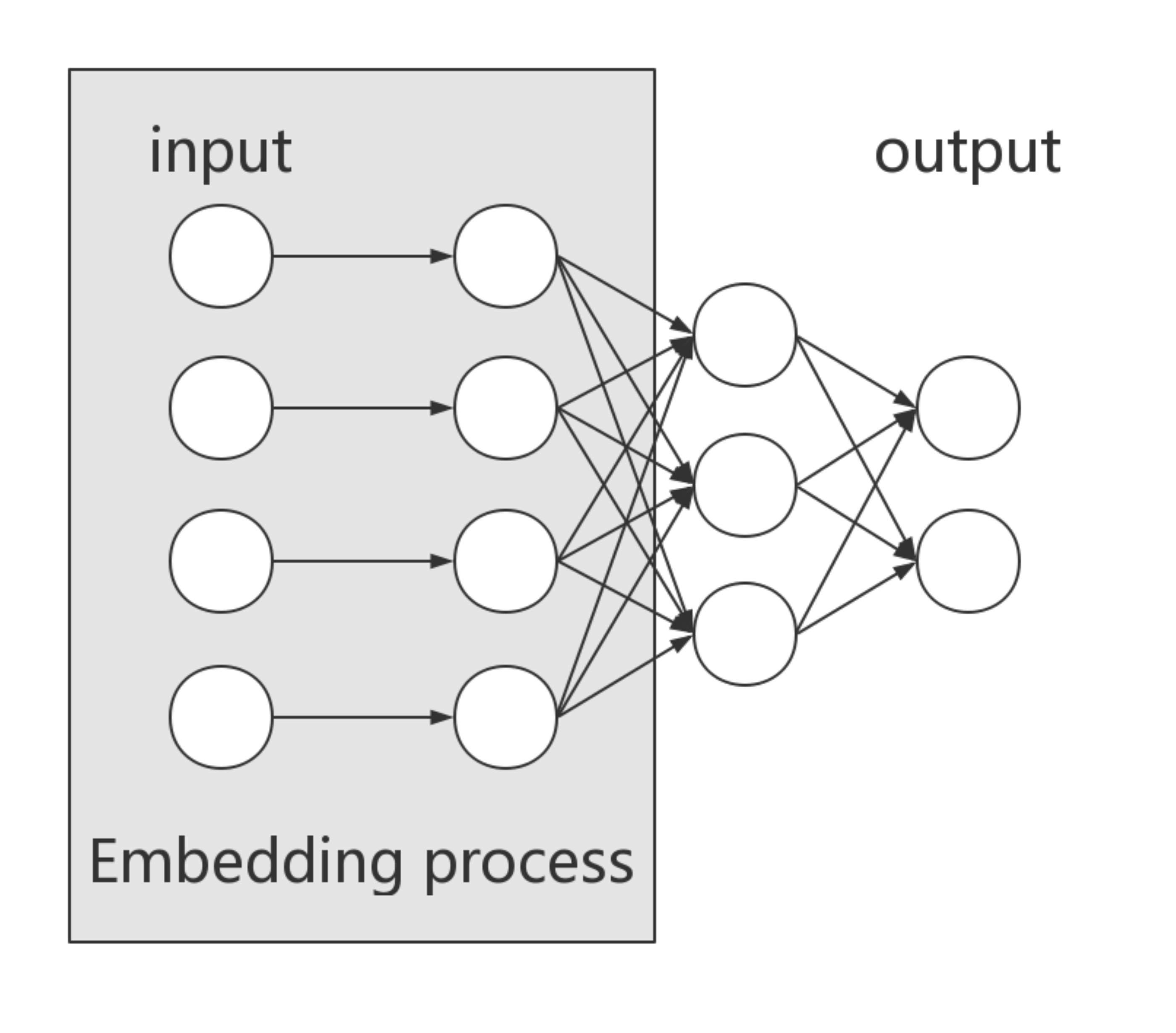}
	\caption{Number learning network (NLN)} \label{fig3}
\end{figure}

\subsection{ Label Decision}
Finally, we get the label number probability for each sample, and choose the label number with the largest value as the final label number. Then we set the value of the label number is $n$, and select the top $n$  of the largest value in the corresponding sample output probability as the final output $R_i$

\section{Experiment}
In this section, we evaluate our proposed approach on the dataset. We introduce the dataset, evaluation measure, experimental details, and all baseline models. Then, we compare our method with the baselines. Finally, we provide the analysis and discussions of experiment results.

\subsection{Dataset and evaluation measure}

All legal data are from Cail2018~\cite{paper_0} that contains more than 2.6 million criminal cases published by the Supreme People's Court of China. And it is divided small and big dataset. In this paper, we used the small dataset that consist of 154,592 samples for training, 32,508 samples for testing. Each of the samples contains complex legal text description and three label types. There is respectively accusation, relating article, and term of penalty. This paper only uses the accusation (label) that consists of 202 accusations. And each sample may have multiple labels. The details of small dataset are illustrated in Table 1. The ratio of the training dataset to the test dataset is about 5:1, and the number of labels is 202. Among them, the size of the single-label samples in the training dataset is about 4/5, and the size of multi-label samples is 1/5. And the number of samples that label number is 2 is the majority in multi-label samples.

\begin{table}\centering
	\caption{The details of Cail2018 small dataset}\label{tab1}
	\begin{tabular}{cccccc}
		\hline
		Dataset & Training & Valid & Test & Label & \\
		\hline
		Quantity & 154592 & 17131 & 32508 & 202 & \\
		\hline
		Label numer	& 1  & 2  & 3  & 4  & Greater than 4  \\
		\hline
		Quantity in training data & 120475 & 30831 & 2914 & 288 & 96\\
		\hline
	\end{tabular}
\end{table}

Because this dataset is provided by Chinese 'fa yan bei' competition\footnote{http://cail.cipsc.org.cn/}, we follow the evaluation measure of the competition. We respectively use micro-F1 and macro-F1 to evaluate the ability of different models that are employed on the dataset.

\subsection{Baselines}
We compare our proposed methods with the following baselines:

TextCNN~\cite{paper_1}: It uses a variety of kernels to extract the key information of the text vectors, and then map the information to the low-dimensional space through the FC network to obtain the output probability.

CRNN~\cite{paper_2}: It uses CNN before using RNN, and finally output the result through FC network.

DPCNN~\cite{paper_3}: It uses down sampling based on ResNet, get more efficient and widely available global information while effectively obtaining local information.

CNN + attention~\cite{paper_13}: While understanding the text information, it finds out the degree of correlation between the text and the label, and maps it to the label space.

Bi-GRU/Bi-LSTM~\cite{paper_4,paper_5}: It better captures longer distance text dependencies.

\subsection{Experiment Configuration}
In experiment, we simply get the word2vec directly after the legal text segmentation, so that the text is mapped into a 512 dimensions vector and the number of words of input data and knowledge respectively is 400, 85. The convolution kernel is generally 3, and for TextCNN, the convolution is respectively [1, 2, 3, 4, 5]. The output dimension of the RNN is 512, and the parameters of the dropout phase are always set to 0.2. Finally, the first layer of the FC network is 1000 dimensions, and the number of dimensions of the second layer’s output is 202. Both loss functions of deep learning model and NLN are cross entropy while training.

For the multi-label charge prediction task, the number of the multi-label samples is generally less than single-label samples. if the label number of samples is too large, such as 8, the number of samples is less than 0.01\% compared to the number of total samples. Therefore the samples are ignored directly by us. Because the number of samples with the number of labels greater than 4 is too small in table 1, we recognized the label number of these samples as 4 in the experiment.  

\subsection{Experiment on Multi-Label Samples}

To prove the effect of our proposed approach, we extract all the samples that the numbers of labels are greater than 1 in the test dataset. We hope to prove the effect of our proposed approach with NLN. The details of multi-label samples are illustrated in Table 2 and the details of experiment are illustrated in Table 3. According to the Table 3, we can clearly see that scores of all the baselines with the number learning network (NLN) are higher in multi-label samples than the ones by threshold strategy. And the improvements of the scores of various baselines are very high.

\begin{table}[!htbp]
	\centering
	\caption{The details of multi-label samples}\label{tab4}
	\begin{tabular}{ccccc}
		Label number & 2 & 3 & 4 & Greater than 4 \\
		\hline
		Quantity & 1913 & 135 & 6 & 2\\
		\hline
	\end{tabular}
\end{table}

With macro-F1 as the index, score of the best model TextCNN is 1.5\% higher than the model with threshold strategy. And DPCNN has the most promotion in the baselines, the score of promotion is 5\%. Meanwhile, we are surprised to find that with micro-F1 as the index, scores of all the baseline model have great improvements. Score of the best model TextCNN is 5\% higher than the model with threshold strategy. The most promotion in the baseline model is DPCNN and the promotion of score is 17\%. Therefore we can get a conclusion that by adding number learning network (NLN), we can make these kinds of deep learning models perform better on multi-label charge prediction task.

\begin{table}[!htbp]
	\centering
	\caption{The details of multi-label samples}\label{tab5}
	\begin{tabular}{cccc}
		\hline
		&  Add NLN & micro—F1(\%) & macro—F1(\%) \\
		\hline
		\multirow{2}*{TextCNN} & with & \textbf{82.47} & \textbf{63.91}\\
		& without & 77.39	& 62.52\\
		\hline
		\multirow{2}*{CRNN} & with & \textbf{78.44} & \textbf{57.97}\\
		& without & 63.90 & 54.63\\
		\hline
		\multirow{2}*{DPCNN} & with & \textbf{80.20} & \textbf{60.80}\\
		& without & 63.90 & 54.63\\
		\hline
		\multirow{2}*{Attention} & with & \textbf{81.81} & \textbf{62.14}\\
		& without & 70.39 & 57.59\\
		\hline
		\multirow{2}*{GRU} & with & \textbf{76.50} & \textbf{53.62}\\
		& without & 60.48 & 50.54\\
		\hline
		\multirow{2}*{LSTM} & with & \textbf{76.62} & \textbf{51.41}\\
		& without & 63.14 & 48.49\\
		\hline
		
	\end{tabular}
\end{table}

\subsection{Comparations of Baselines }
In this part, we apply baselines to the whole dataset including single label samples and multi-label samples.  The details of experiment results are illustrated in the Table 4. In Table 4, the single-label type shows that we only select the label with the highest probability as final prediction result, so we recognize it as single-label task; the multi-label type shows that we have selected all the labels that meet the conditions by threshold strategy as the output. Here we use the appropriate values as the threshold by many manual selections. We find that in all the baselines, using the RNN model to process word2vec vectors that the fact converted is the worst, and using TextCNN model with five different convolution kernels performs best. With micro-F1 as the evaluation index, the best model TextCNN scores 10\% higher than the worst model Bi-GRU. With macro-F1 as the evaluation index, the best model TextCNN scores 15\% higher than the worst model Bi-GRU.  

Although the data is extremely imbalance, we can still find that after the selected threshold determining results, the F1 scores have improved, but the effect is not satisfactory. On this dataset, the difference between the multi-label results with all baselines and the single-label results is not large. Based on the evaluation measure, their improvement is less than 1\% on average. Therefore we can make a judgment that the widely threshold strategy has defects in accuracy.

\begin{table}[!htbp]
	\centering
	\caption{Results of baselines}\label{tab2}
	\begin{tabular}{cccc}
		\hline
		 &  type & micro—F1(\%) & macro—F1(\%) \\
		\hline
		\multirow{2}*{TextCNN} & single-label & 84.59 & 74.54\\
		& multi-label & \textbf{85.95} & \textbf{76.18}\\
		\hline
		\multirow{2}*{CRNN} & single-label & 77.67	& 61.19\\
		& multi-label & 78.04 & 61.58\\
		\hline
		\multirow{2}*{DPCNN} & single-label & 80.55 & 65.03\\
		& multi-label & 80.9 & 65.49\\
		\hline
		\multirow{2}*{Attention} & single-label & 82.69 & 71.36\\
		& multi-label & 83.44 & 72.10\\
		\hline
		\multirow{2}*{GRU} & single-label & 75.48 & 58.16\\
		& multi-label & 75.76 & 58.44\\
		\hline
		\multirow{2}*{LSTM} & single-label & 77.54 & 60.17\\
		& multi-label & 77.96 & 62.20\\
		\hline
		
	\end{tabular}
\end{table}

\subsection{Results of Our Approach}

In this part, we use our purposed approach with all the baseline models. The details of experiment are illustrated in Table 5. In Figure 5, we find that our approach has some improvements on various baselines. According to the Table 5, with micro-F1 as the index, we can observe that the results using our approach have a better performance. For all models with our approach the scores increase, and some models have 5\%-8\% improvements; With macro-F1 as the index, we can also observe that all the models have great improvements, and some models have 9\%-13\% improvements. Scores of most models reach above 70\%. Because of the definition of macro-F1, we can also make a conclusion: we can correctly classify more multi-label samples with our approach. And results prove that our approach can solve this task better.

\begin{table}[!htbp]
	\centering
	\caption{Results of baselines with our approach}\label{tab3}
	\begin{tabular}{cccc}
		\hline
		&  Our Approach & micro—F1(\%) & macro—F1(\%) \\
		\hline
		\multirow{2}*{TextCNN} & True & \textbf{86.23} & \textbf{77.50}\\
		& Flase & 85.95 & 76.18\\
		\hline
		\multirow{2}*{CRNN} & True & \textbf{84.58} & \textbf{71.83}\\
		& Flase & 78.04 & 61.58\\
		\hline
		\multirow{2}*{DPCNN} & True & \textbf{85.74} & \textbf{76.06}\\
		& Flase & 80.9 & 65.49\\
		\hline
		\multirow{2}*{Attention} & True & \textbf{83.50} & \textbf{74.52}\\
		& Flase & 83.44 & 72.10\\
		\hline
		\multirow{2}*{GRU} & True & \textbf{83.69} & \textbf{71.63}\\
		& Flase & 75.76 & 58.44\\
		\hline
		\multirow{2}*{LSTM} & True & \textbf{78.96} & \textbf{61.85}\\
		& Flase & 77.98 & 60.66\\
		\hline
		
	\end{tabular}
\end{table}

\section{Conclusions and Future Work}
n this paper, we propose an external knowledge enhanced end-to-end multi-label charge prediction approach with automatic label number learning and a number learning network (NLN). By using it, we prove that the approach can improve the performance of the deep learning model in the multi-label charge prediction task. This method has great improvement on all models. At the same time, we extract all the multi-label samples from the test dataset and test the approach on it. The experiment results show that our approach is better than the deep learning model with the threshold strategy. With macro-F1 as the index, the improvement of various deep learning models is 3\%-5\%; With micro-F1 as the index, it has increased 5\%-15\%. 

In the future, we will connect deep learning models with number learning network (NLN) while training, rather than getting the output probability of sample first, and create a new loss function that is used in training. We hope that can improve accuracy of the charge prediction and reduce the training loss error. In particularly, we want to avoid predicting the single-label samples into multi-label category to increase the score of F1. At the same time, we will test the performance of NLN on other multi-label tasks, and hope that further study can improve its versatility.

%
%
%
%

\bibliographystyle{splncs04}
\bibliography{reference}

\end{document}